\pdfoutput=1

\documentclass[11pt]{article}

\usepackage[]{acl}

\usepackage{times}
\usepackage{latexsym}
\usepackage{booktabs}
\usepackage{algpseudocode}
\usepackage{tikz}
\usepackage{tikz-qtree}
\usepackage{amsfonts}
\usepackage{enumitem}

\usepackage[T1]{fontenc}

\usepackage[utf8]{inputenc}

\usepackage{microtype}
\usepackage{inconsolata}

\usepackage{relsize}
\usepackage{xspace}

\newcommand{\library}{\texttt{string2string}\xspace}

\DeclareRobustCommand{\librarylogo}{%
  \begingroup\normalfont
  \vspace{-0.2em}%
  \raisebox{-0.4em}{%
  \includegraphics[height=3.5em]{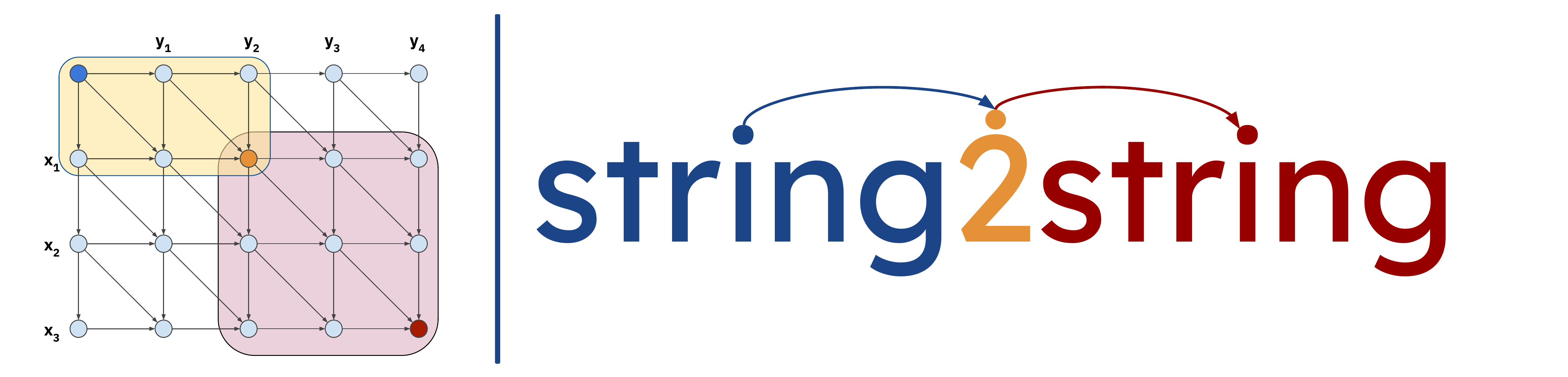}%
  }%
  \vspace{0.2em}
  \kern 0.4em%
  \endgroup
}

\usepackage[labelfont=bf]{caption}

\title{
\librarylogo\\
\mbox{\library: A Modern Python Library for String-to-String Algorithms}
}

\author{
Mirac Suzgun\\
Stanford University \\
\And
Stuart M. Shieber\\
Harvard University \\ \\ 
\emph{\normalsize{In memory of Lynn.}}
\And
Dan Jurafsky\\
Stanford University \\
}

\begin{document}

 \maketitle
\begin{abstract}
We introduce \textbf{\library}, an open-source library that offers a comprehensive suite of efficient algorithms for a broad range of string-to-string problems. It includes traditional algorithmic solutions as well as recent advanced neural approaches to tackle various problems in string alignment, distance measurement, lexical and semantic search, and similarity analysis---along with several helpful visualization tools and metrics to facilitate the interpretation and analysis of these methods. Notable algorithms featured in the library include the Smith-Waterman algorithm for pairwise local alignment, the Hirschberg algorithm for global alignment, the Wagner-Fisher algorithm for edit distance, BARTScore and BERTScore for similarity analysis, the Knuth-Morris-Pratt algorithm for lexical search, and Faiss for semantic search. Besides, it wraps existing efficient and widely-used implementations of certain frameworks and metrics, such as sacreBLEU and ROUGE, whenever it is appropriate and suitable. Overall, the library aims to provide extensive coverage and increased flexibility in comparison to existing libraries for strings. It can be used for many downstream applications, tasks, and problems in natural-language processing, bioinformatics, and computational social sciences. 
It is implemented in Python, easily installable via pip, and accessible through a simple API. 
Source code, documentation, and tutorials are all available on our GitHub page: \url{https://github.com/stanfordnlp/string2string}.\footnote{Correspondence to: \url{msuzgun@cs.stanford.edu}.}
\end{abstract}

\vspace{-0.9em}
\section{Introduction}
\vspace{-0.5em}

String-to-string problems have a wide range of applications in various domains and fields, including, but not limited to, natural-language processing (e.g., information extraction, spell checking, and semantic search), computational molecular biology (e.g., DNA sequence alignment), programming languages and compilers (e.g., parsing and compiling), as well as computational social sciences and  digital humanities (e.g., lexical and semantic analysis of literary texts and corpora). 

The current state of string-to-string processing, alignment, distance, similarity, and search algorithms is marked by a multitude of implementations in widely used programming languages, such as C++, Java, and Python. However, many of these implementations are not integrated with one another---they also lack flexibility, modularity, and comprehensive documentation, hindering their accessibility to users. As such, there is a pressing need for a unified platform that combines these functionalities into one accessible and comprehensive system.

In this work, we present an open-source library that offers a broad collection of algorithms and techniques for the alignment, manipulation, or evaluation of string-to-string mappings.\footnote{We define a \emph{string} as an ordered collection of characters---such as letters, numerals, symbols---which serves as a representation of a unit of information, text, or data. Strings can be used to represent anything, from simple sentences to complex nucleic acid sequences or elaborate computer programs.} These problems include measuring the lexical distance between two strings (e.g., under the Levenshtein edit distance metric), computing the local or global alignment between two DNA sequences (e.g., based on a substitution matrix such as BLOSUM), calculating the semantic similarity between two texts (e.g., using BART-embeddings), and performing efficient semantic search (e.g., via the Faiss library by FAIR~\citep{johnson2019billion}).

The \library library has been purposefully crafted to prioritize key design principles, including modularity, completeness, efficiency, flexibility, and clarity. 
As an open-source initiative, the library will continue to grow and adapt to meet the  evolving of its user community in the future, and we are committed to ensuring that the library remains a flexible, accessible, and dynamic resource, capable of accommodating the changing landscape of string-to-string problems and tasks.

\begin{figure*}[!t]
\centering
\includegraphics[width=1.0\textwidth]{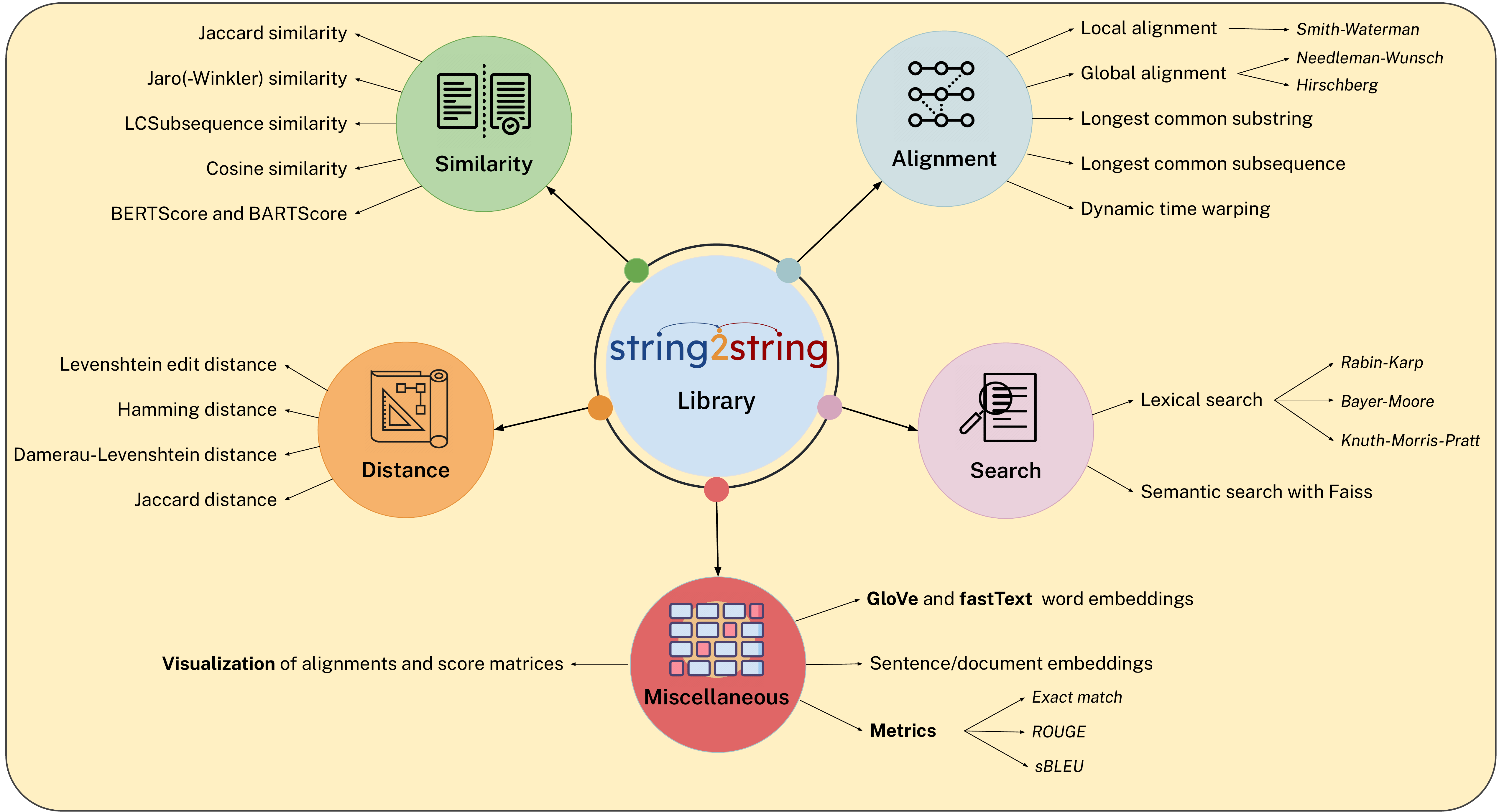}
\caption{The \library library provides a broad set of algorithms and techniques to tackle a variety of problems and tasks involving the pairwise alignment, comparison, evaluation, manipulation, and processing of string-to-string mappings. It includes the implementation of widely-used algorithms, such as Smith-Waterman~\citep{SmithWaterman1981} for local alignment, Knuth-Morris-Pratt~\citep{knuthmorrispratt1977} for identical string matching (search), and Wagner-Fisher~\citep{wagner1974string} for edit distance, as well as recent neural approaches, such as BERTScore~\citep{bertscore2020} and BARTScore~\citep{bartscore2021} for semantic similarity measurements and Faiss~\citep{johnson2019billion} for semantic search. The library has been designed to support not only individual strings but also lists of strings so that users can align and compare strings at the token, word, or sentence levels. It further contains visualization features to allow users to visualize alignments and score matrices of strings. 
}
\label{fig:overview}
\end{figure*}

\section{Related Work}
The fields of natural-language processing and machine learning have a long-standing and exemplary tradition of fostering a culture that values open-source tools and libraries. While designing our own string-to-string library, which includes both traditional algorithmic and neural approaches to various problems, we have drawn inspirations from 
Natural Language Toolkit (NLTK; \citet{bird-loper-2004-nltk}), 
Gensim \citep{rehurek_lrec-2010}, 
OpenGrm Ngram \citep{roark-etal-2012-opengrm},
Stanford CoreNLP Toolkit \citep{manning-etal-2014-stanford},
OpenNMT \citep{klein-etal-2017-opennmt},
tensor2tensor \citep{waswani-et-al-2018-tensor2tensor}, 
AllenNLP \citep{gardner-etal-2018-allennlp}, 
fairseq \citep{ott-etal-2019-fairseq},
spaCy \citep{neumann-etal-2019-scispacy}, 
Stanza \citep{qi-etal-2020-stanza}, 
Transformers \citep{wolf-etal-2020-transformers}, and 
Torch-Struct \citep{rush-2020-torch},
among many others.

\begin{table*}[!t]
\small 
\centering
\hspace*{-0.3cm}\begin{tabular}{p{165mm}} %
  \toprule
  \multicolumn{1}{c}{\colorbox[HTML]{b9e9e3}{\textbf{Pairwise Alignment}}} \\
    \midrule
    \colorbox[HTML]{b9e9e3}{\textbf{Local alignment}}: The best possible matching substring or subsequence alignment between two strings---based on a substitution matrix and  gap penalty function---allowing for gaps and mismatches within a specified region of the input sequences. \\
    $\star$ (a) Dynamic programming solution \citep{SmithWaterman1981}: $\mathcal{O}(nm)$ in terms of space and time. \\
    \midrule
    \colorbox[HTML]{b9e9e3}{\textbf{Global alignment}}: The best possible alignment between two strings over their entire length. \\
    $\star$ (a) Dynamic programming solution \citep{needleman1970general}: $\mathcal{O}(nm)$ in terms of space and time. \\
    $\star$ (b) Divide-and-conquer + dynamic programming solution \citep{hirschberg1975linear}: $\mathcal{O}(m)$ in terms of space and $\mathcal{O}(nm)$ time. \\
    \midrule
    \colorbox[HTML]{b9e9e3}{\textbf{Longest common substring}}: The longest \emph{contiguous} substring that appears in both strings.\\
    $\star$ (a) Dynamic programming solution: $\mathcal{O}(nm)$ in terms of space and time. \\
    \midrule
    \colorbox[HTML]{b9e9e3}{\textbf{Longest common subsequence}}: The longest possible sequence of symbols that appears in the same order in both strings. \\
    $\star$ (a) Dynamic programming solution: $\mathcal{O}(nm)$ in terms of space and time. \\
    \midrule
    \colorbox[HTML]{b9e9e3}{\textbf{Dynamic time warping (DTW)}}: The optimal warp path that minimizes the distance between sequences of varying length. \\
    $\star$ (a) Dynamic programming solution \citep{sakoe1978dynamic}: $\mathcal{O}(nm)$ in terms of space and time. \\
    $\star$ (b) Space-time improved version of (a) via \citet{hirschberg1975linear}'s algorithm: Reduces space complexity to $\mathcal{O}(m)$.\\
    \midrule
  \multicolumn{1}{c}{\colorbox[HTML]{fdd55b}{\textbf{Distance}}} \\ %
    \midrule
    \colorbox[HTML]{fdd55b}{\textbf{Levenshtein edit distance}}: The minimum number of insertions, deletions, and substitutions needed to convert $S$ into $T$. \\
    $\star$ (a) Dynamic programming solution  \citep{wagner1974string}: $\mathcal{O} (nm)$ in terms of space and time.\\ 
    $\star$ (b) Space-improved version of (a): Reduces space complexity to $\mathcal{O}(m)$ by storing only two rows.\\
    \midrule
    \colorbox[HTML]{fdd55b}{\textbf{Hamming distance}}: The total number of indices at which strings, $S$ and $T$, of equal length differ. \\
    $\star$ (a) Naive solution: $\mathcal{O}(n)$ in terms of space and time. \\
    \midrule
    \colorbox[HTML]{fdd55b}{\textbf{Damerau–Levenshtein distance}}: The minimum number of insertions, deletions,  substitutions, and adjacent transpositions needed to convert $S$ into $T$. \\
    $\star$ (a) Dynamic programming solution (simple extension of the Wagner-Fisher algorithm): $\mathcal{O} (nm)$ in terms of space and time.\\
    $\star$ (b) Space-improved version of (a): Reduces space complexity to $\mathcal{O}(m)$ by storing only two rows.\\
    \midrule
    \colorbox[HTML]{fdd55b}{\textbf{Jaccard distance}}: The inverse of Jaccard similarity (that is, 1.0 - Jaccard similarity coefficient). \\
    $\star$ (a) Naive solution: $\mathcal{O}(n)$ in terms of space and time. \\
    \bottomrule
    \end{tabular}
    \caption{Overview of the pairwise string alignment and distance problems addressed by the library, along with the algorithmic approaches employed to solve them. In all instances, we assume that we are given two strings, $S$ and $T$, over a finite alphabet $\Sigma$, where where $n=|S|$, $m=|T|$, and $k=|\Sigma|$, with $m \leq n$. Also, whenever possible, we include the brute-force and memoized solutions to these problems (as in the case of edit distance, for instance).}
    \label{tab:alignment-distance}
\end{table*}

\section{Overview of Algorithms}
The \library library offers a rich collection of algorithmic solutions to tackle a wide range of string-to-string problems and tasks. We have clustered these algorithms into four categories: pairwise alignment, distance measurement, similarity analysis, and search.\footnote{By duality, distance measurement methods can naturally be used for string similarity analysis, and vice versa.} Each category contains a suite of efficient algorithms that are tailored to address specific problems within their respective domain. In what follows, we provide a brief overview of these algorithms, along with the associated problems or tasks they are designed to solve.

\begin{table*}[!t]
\small 
\centering
\hspace*{-0.3cm}\begin{tabular}{p{165mm}} %
  \toprule
  \multicolumn{1}{c}{\colorbox[HTML]{d3ff9f}{\textbf{Similarity}}} \\
    \midrule
    \colorbox[HTML]{d3ff9f}{\textbf{Jaccard similarity}}: The size of the set of unique symbols that appear in both strings (i.e., in the intersection) divided by the size of the set of the union of the symbols in both strings.\\
    $\star$ (a) Naive solution: $\mathcal{O}(n)$ in terms of space and time. \\   
    \colorbox[HTML]{d3ff9f}{\textbf{Jaro(-Winkler) similarity}}: A measure of similarity based on matching symbols and transpositions in two strings. \\
    $\star$ (a) Naive solution: $\mathcal{O} (nm)$ in terms of time and $\mathcal{O} (n)$ in terms of space. \\  
    \colorbox[HTML]{d3ff9f}{\textbf{LCSubsequence similarity}}: A degree of similarity between two strings based on the length of their longest common subsequence. \\
    $\star$ (a) Based on the efficient solution to the longest common subsequence problem.{\textsuperscript{$\clubsuit$}} \\
    \colorbox[HTML]{d3ff9f}{\textbf{Cosine similarity}}: The similarity between two strings based on the angle between their corresponding vector representations.\\
    $\star$ (a) Utilizes \texttt{numpy} and \texttt{torch} functions: $\mathcal{O} (E)$ in terms of time and $\mathcal{O} (E)$ in terms of space.{\textsuperscript{$\diamondsuit$}} \\
    \colorbox[HTML]{d3ff9f}{\textbf{BERTScore}} \citep{bertscore2020}: A measure of semantic similarity that employs contextualized embeddings derived from the pre-trained BERT model \citep{devlin-etal-2019-bert} to estimate the semantic closeness between two pieces of text. \\
    $\star$ (a) Adaptation of the original BERTScore implementation: $\mathcal{O} (nm)$ in terms of time and $\mathcal{O} (nm \cdot E)$ in terms of space.\\
    \colorbox[HTML]{d3ff9f}{\textbf{BARTScore}} \citep{bartscore2021}: A measure of semantic similarity that utilizes the pre-trained BART model \cite{lewis2020bart} and that achieves high correlation with human judgements. \\
     $\star$ (a) Adaptation of the original BARTScore implementation: $\mathcal{O} (nm)$ in terms of time and $\mathcal{O} (nm \cdot E)$ in terms of space.\\
    \midrule
  \multicolumn{1}{c}{\colorbox[HTML]{fbbad7}{\textbf{Search}}} \\ %
    \midrule
    \colorbox[HTML]{fbbad7}{\textbf{Lexical search}}:   \\
    $\star$ (a) Naive (brute-force) search: $\mathcal{O}(mn)$ in terms of match time and $\mathcal{O}(1)$ in terms of space. \\
    $\star$ (b) Rabin-Karp algorithm \citep{karp1987efficient}: $\mathcal{O}(mn)$ in terms of match time and $\mathcal{O}(1)$ in terms of space. \\
    $\star$ (c) Boyer-Moore algorithm \citep{boyermoore1977}: $\mathcal{O}(mn)$ in terms of match time and $\mathcal{O}(|\Sigma|)$ in terms of space.. \\
    $\star$ (d) Knuth-Morris-Pratt algorithm \citep{knuthmorrispratt1977}: $\mathcal{O}(n)$ in terms of match time and $\mathcal{O}(m)$ in terms of space. \\
    \colorbox[HTML]{fbbad7}{\textbf{Semantic search}}:   \\
    $\star$ (a) FAISS~\citep{johnson2019billion}: $\mathcal{O}(\log^2 n)$ in terms of match time and $\mathcal{O}(n \cdot E)$ in terms of space.{\textsuperscript{$\spadesuit$}}\\
    \bottomrule
    \end{tabular}
    \vspace{-0.2em}
    \caption{Overview of the string similarity and search solutions used in the library. As in Table~\ref{tab:alignment-distance}, we assume that we are provided with two strings, $S$ and $T$, over a finite alphabet $\Sigma$, where $n=|S|$, $m=|T|$, and $k=|\Sigma|$, with $m \leq n$. Furthermore, we use $E$ to denote the size of the embedding space (or token), whenever applicable.
    Both the Rabin-Karp and Knuth-Morris-Pratt algorithms require $\Theta(m)$ time for pre-processing and $\Theta(n)$ time for searching, whereas the Boyer-Moore algorithm has a pre-processing time complexity of $\Theta(m+k)$. In terms of space, the Rabin-Karp, Boyer-Moore, and Knuth-Morris-Pratt algorithms require $\Theta(1)$, $\Theta(k)$. and $\Theta(m)$, respectively.
    Footnote {\textsuperscript{$\clubsuit$}}: Please refer to Eqn. (11) in \citep{suzgun2022crowdsampling} for a mathematical formulation of LCSubsequence similarity. Note that the authors call this similarity measure ``\texttt{Sim-LCS}.'' Footnote {\textsuperscript{$\diamondsuit$}}: We assume that the dimension (size) of the two vectors are both $E$. Footnote {\textsuperscript{$\spadesuit$}}: We invite our readers to look at the blogpost by \citet{feinberg2019} for a detailed analysis of Facebook AI Research's Faiss algorithm.}
    \label{tab:similarity-search}
\end{table*}

\subsection{Pairwise Alignment}
Pairwise string alignment is the problem of identifying an optimal alignment between two strings, such as nucleotide sequences in DNA or paragraphs in a text. This task involves aligning them in such a way that maximizes the number of matching symbols while allowing for gaps or mismatches where necessary. Pairwise string alignment is a widely-used technique that plays a crucial role in tasks such as DNA sequence alignment, database searching, and phylogenetic analysis. 

As exhibited in Table~\ref{tab:alignment-distance}, the library, in its current state, provides efficient solutions to local alignment, global alignment, longest common substring (LCSubstring), longest common subsequence (LCSubsequence), and dynamic time warping (DTW) problems. It is worth noting that all of the problems and tasks covered in this suite can be solved using standard dynamic programming-based solutions. Alternative approaches to long sequence or string alignment problems, such as FASTA~\citep{lipman1985rapid} and  BLAST~\citep{altschul1990basic}, also exist; they offer improved efficiency through the use of probabilistic or heuristic methods, but they do not always guarantee optimal solutions and may sacrifice accuracy for speed. Due to their ability to handle large datasets quickly and provide reasonably accurate results, BLAST and FASTA are still widely used in bioinformatics, and for that reason, we plan on including them in the \library library in the  future.

\subsection{Distance}
String distance refers to the problem of quantifying the extent to which two given strings are dissimilar based on a distance function. The Levenshtein edit distance metric, for instance, corresponds to the minimum number of insertion, deletion, or substitution operations required to transform one string into another. It has a famous dynamic programming solution, which is often referred to as the Wagner-Fisher algorithm \citep{wagner1974string}. In this library, we provide an implementation of the Wagner-Fisher algorithm---which has a quadratic time and space complexity, as well as an improved version of it---which reduces the overall space complexity to linear.\footnote{Incidentally, we highlight an important discovery by \citet{backursindyk2015} that the edit distance between two strings cannot be computed in strongly subquadratic time, unless the strong exponential time hypothesis is false.} We further cover and provide efficient solutions to the Hamming distance, Damerau-Levenshtein distance, and Jaccard distance problems, as shown in Table~\ref{tab:alignment-distance}.

One noteworthy feature of the library is that it allows the user to specify the weight of string operations (viz., insertions, deletions, substitutions, and transpositions) depending on the distance function of choice. Furthermore, it can compute the distance between not only string pairs but also pairs of lists of strings, thereby not limiting the users to make comparisons only at the character or symbol level.

\subsection{Similarity}
String similarity refers to the problem of measuring the degree to which two given strings are similar to each other based on a similarity function---which can be defined on various criteria, such as character matching, longest common substring or subsequence comparison, or structural alignment. There is a natural duality between string similarity measures and string distance measures, which means that it possible to convert one into the other with ease; hence, it is often the case that one uses string similarity and distance measures interchangeably.

Jaccard similarity, Jaro similarity, Jaro-Winkler similarity, LCSubsequence similarity, cosine similarity, BERTScore, and BARTScore are among the similarity measures that are covered in the library. The first four can be seen as lexical similarity measures, as they assess surface or structural closeness, whereas the remaining three can be regarded as semantic similarity measures, as they incorporate the implied meaning of the constituents of the given strings into account.

The present library provides users with the ability to calculate cosine similarity not only between individual words---via pre-trained GloVe \citep{pennington-etal-2014-glove} or fastText \citep{joulin2016fasttext} word embeddings---but also for longer pieces of text such as sentences, paragraphs, or even documents---via averaged or last-token embeddings obtained from a neural language model such as BERT \citep{devlin-etal-2019-bert}. As we also mention in Section~\ref{sec: additional-features}, this feature enables users to compare the semantic similarity of longer segments of text in only a few lines of code, providing greater flexibility in text analysis tasks. 

\subsection{Search}
String search, also known as string matching, refers to the problem of determining whether a given pattern string exists inside a longer string. The brute-force approach to string search would involve examining each position of the longer string to determine if it matches the pattern string; however, this method can be inefficient, particularly when dealing with large strings. In the library, we therefore include the Rabin-Karp \citep{karp1987efficient}, Boyer-Moore \citep{boyermoore1977}, and Knuth-Morris-Pratt \citep{knuthmorrispratt1977} algorithms for identical string matching as well.

The library additionally provides support for semantic search via Facebook AI Research's Faiss library \citep{johnson2019billion}, which, in essence, allows efficient similarity search and clustering of dense vectors. In contrast to the previous setup for identical string matching, Faiss initially requires the user to provide a \emph{list} of strings (texts) as a corpus and creates a fixed-vector representation of each string using a neural language model.\footnote{The user is provided with the flexibility to determine how to obtain a fixed embedding for each text. Specifically, the user has the option to choose between different embedding methods, such as averaging the token embeddings or selecting the embedding of the final token in the sequence.} Once the initialization of the corpus is done, one can perform ``queries'' on the corpus. Given a new query, one can automatically get the embedding of that query, map it onto the embedding space of the corpus, and return the nearest neighbours of the query on the embedding space, thereby finding the texts that are semantically closest to the query.  

As one might imagine, string search algorithms are highly practical tools that have a wide range of applications across different fields. These algorithms allow users to locate and retrieve specific patterns within a long text or a large corpus. For instance, the string search algorithms covered in the \library library---as shown in Table~\ref{tab:similarity-search}---can be used for pattern recognition, DNA matching, plagiarism detection, and data mining, among many other downstream applications and tasks.

\section{Additional Features}
\label{sec: additional-features}
\vspace{-0.3em}

One noteworthy feature of the library is its ability to simplify the use of the GloVe~\citep{pennington-etal-2014-glove} or fastText~\cite{joulin2016fasttext} word embeddings by enabling users to download and use them with just one line of code. This streamlined process not only saves users time and effort but also eliminates the need for additional installations or complex configurations. By providing this feature, we have sought to make the library more accessible to users and encourage the use of pre-trained word embeddings in various string-to-string tasks and applications, such as measuring the cosine similarity between two words.\footnote{Notably, fastText offers pre-trained word embeddings for 157 languages---trained on Common Crawl and Wikipedia---that one can easily download and use with our library.}

Similarly, users can seamlessly get the averaged or last token embeddings of a piece of text from a pre-trained language model that is hosted on Hugging Face Models
~\citep{wolf-etal-2020-transformers} or on their local path in a few lines of code, and we provide both CPU and GPU support for these computations.

\begin{figure}[t]
\centering
\vspace{-0.8em}
\includegraphics[width=1.05\columnwidth]{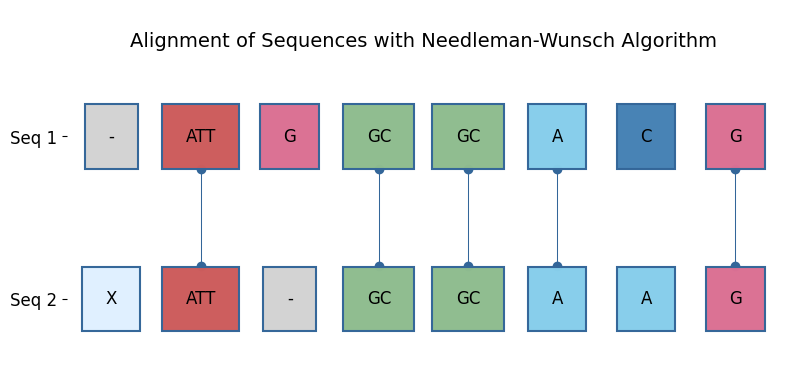}
\vspace{-2.7em}
\caption{Alignment of two sequences, \texttt{[ATT G GC GC A C G]} and \texttt{[X ATT GC GC A A G]}, as obtained by the Needleman-Wunsch algorithm \citep{needleman1970general}. Our library allows users to visualize the pairwise alignment between strings (or lists of strings).}
\label{fig:seqn-algn-ex}
\vspace{-0.3em}
\end{figure}

Finally, we note that the library offers various visualization capabilities that allow users to visually inspect the alignment between two strings or the score matrix of the distance or similarity between them. This functionality facilitates the understanding and interpretation of the output of various algorithms and can aid in the selection of the most suitable algorithm for a given task. By incorporating this feature into the library, we aim to enhance the user experience and provide intuitive means of interpreting the outputs.\footnote{For instance, our library provides a practical, hands-on tutorial focused on the HUPD \citep{suzgun2022harvard}, a large patent corpus. This tutorial showcases the efficient use of our library's functionalities and features for performing semantic search and visualizing the textual content of patent documents.} Figure~\ref{fig:seqn-algn-ex} shows a simple alignment between two lists of strings, as generated by our library.

\section{Library Design Principles}
\vspace{-0.5em}

We have endeavoured to build a comprehensive and easy-to-use platform for numerous string-to-string processing, comparison, manipulation, and search algorithms. We have purposefully structured and organized the library to allow easy customization, functional extension, and modular integration. 

\textbf{\emph{Completeness.}} The library offers a comprehensive set of classical algorithms as well as neural approaches to tackle a wide range of string-to-string problems. We have intentionally included both efficient and simpler solutions, such as brute-force and memoization-based approaches, where appropriate. By providing multiple solutions to the same problem, the library allows users to compare and contrast the performance of different algorithmic methods. This approach enhances users' understanding of the trade-offs between different algorithmic solutions and helps them appreciate the strengths and limitations of each approach.

\textbf{\emph{Modularity.}} To improve the efficiency and maintainability of the codebase, we have adopted a modular design approach that breaks down the code into smaller and self-contained modules. This approach simplifies the process of adding new features and functionalities and modifying existing ones for developers, while also enabling efficient testing, debugging, and overall maintenance of the library. The modular design has allowed us to quickly locate and fix any errors, without disrupting the entire codebase, during development. Moreover, this modular approach ensures that the library is scalable and adaptable to future updates and changes, which should enable us to easily improve the library's functionality and expand its use in various tasks and applications moving forward.

\textbf{\emph{Efficiency.}} We have taken great care to ensure that the algorithm implementations are efficient both computationally and memory-wise so that they could easily handle large datasets and complex tasks. We provide basic support for process-based parallelism via Python's inherent \href{https://docs.python.org/3/library/multiprocessing.html}{\texttt{multiprocessing}} package, as well as \href{https://joblib.readthedocs.io/en/latest/}{\texttt{joblib}}. 
Additionally, we provide GPU support for neural-based approaches, whenever applicable. While we strive to balance efficiency and clarity, we acknowledge that in some cases, trade-offs may exist between the two. In such cases, we have placed greater emphasis on clarity, ensuring that the algorithms are transparent and easy to understand, even at the cost of some efficiency. Nonetheless, we believe that the library's overall efficiency, combined with its transparency and comprehensibility, makes it a valuable resource for the community.

\textbf{\emph{Support for List of Strings.}} The library has been designed to support not only individual strings but also lists of strings---whenever possible, enabling users to align or compare strings at the subword or token level. This feature provides greater flexibility in the library's use cases, as it allows users to analyze and compare more complex data structures beyond only individual strings. By supporting lists of strings, the library can handle a wider range of textual input types and structures.

\textbf{\emph{Strong Typing.}} The use of strong typing requirements is an essential aspect of the library, as it ensures that the inputs are always consistent and accurate, which is crucial for generating reliable results. By carefully annotating all the arguments of the algorithms used in the library, we have sought to increase the robustness and reliability of the codebase. This approach has helped prevent input-related errors, such as incorrect data type or format, from occurring during execution.

\textbf{\emph{Accessibility.}} The library has been implemented in Python, a programming language which has been the core of many natural-language processing tools and applications in academia and industry. The \library library is ``pip''-installable and can be integrated into common machine learning and natural-language processing frameworks such as PyTorch, TensorFlow, and scikit-learn.

\textbf{\emph{Open-Source Effort.}} The library is---and will remain---free and accessible to all users. We hope that this approach will promote community-driven development and encourage collaboration among researchers and developers, enabling them to contribute to and improve the library.

\section{Conclusion}

We introduced \library, an open-source library that offers a large collection of algorithms for a broad range of string-to-string problems. The library is implemented in Python, hosted on GitHub, and installable via pip. It contains extensive documentation along with several hands-on tutorials to aid users to explore and utilize the library effectively. With the help of the open-source community, we hope to grow and improve the library. We encourage users to feel free to provide us with feedback, report any issues, and propose new features to expand the functionality and scope of the library. 

\clearpage

\section*{Acknowledgements}
We would like to express our deepest appreciation to Corinna Coupette for providing us with the inspiration to create this open-source library. Furthermore, we would like to thank
Sarah DeMott,
Sebastian Gehrmann,
Elena Glassman,
Tayfun G\"{u}r,
Daniel E. Ho,
\c{S}ule Kahraman,
Deniz Kele\c{s},
Scott D. Kominers,
Christopher Manning,
Luke Melas-Kyriazi,
Tol\'{u}l\d{o}p\d{\'{e}} \`{O}g\'{u}nr\d{\`{e}}m\'{i}, 
Alexander ``Sasha'' Rush,
George Saussy,
Kutay Serova,
Holger Spamann,
Kyle Swanson, and
Garrett Tanzer
for their valuable comments, useful suggestions, and support. 
We owe a special debt of gratitude to Federico Bianchi for inspecting our code, documentation, and tutorials many times and providing us with such helpful and constructive feedback.

\bibliography{anthology,custom}
\bibliographystyle{acl_natbib}

\end{document}